\documentclass{article}

\usepackage{microtype}
\usepackage{graphicx}
\usepackage{subfigure}
\usepackage{booktabs} 

\usepackage{hyperref}

\usepackage[accepted]{icml2025}

\usepackage{amsmath}
\usepackage{amssymb}
\usepackage{mathtools}
\usepackage{amsthm}

\usepackage[capitalize,noabbrev]{cleveref}

\theoremstyle{plain}

\theoremstyle{definition}

\theoremstyle{remark}

\usepackage[textsize=tiny]{todonotes}

\icmltitlerunning{Practical Principles for AI Cost and Compute Accounting}

\begin{document}

\twocolumn[
\icmltitle{Practical Principles for AI Cost and Compute Accounting}

\icmlsetsymbol{equal}{*}

\begin{icmlauthorlist}
\icmlauthor{Stephen Casper}{mit}
\icmlauthor{Luke Bailey}{stanford}
\icmlauthor{Tim Schreier}{fli}
\end{icmlauthorlist}

\icmlaffiliation{mit}{MIT CSAIL (this work was done partly at MIT CSAIL and partly independently)}
\icmlaffiliation{stanford}{Stanford University}
\icmlaffiliation{fli}{Future of Life Institute}

\icmlcorrespondingauthor{Stephen Casper}{scasper@mit.edu}

\icmlkeywords{Machine Learning, ICML}

\vskip 0.3in
]



\printAffiliationsAndNotice{}  

\begin{abstract}

Policymakers increasingly use development cost and compute as proxies for AI capabilities and risks.
Recent laws have introduced regulatory requirements for models or developers that are contingent on specific thresholds. 
However, technical ambiguities in how to perform this accounting create loopholes that can undermine regulatory effectiveness. 
We propose seven principles for designing AI cost and compute accounting standards that (1) reduce opportunities for strategic gaming, (2) avoid disincentivizing responsible risk mitigation, and (3) enable consistent implementation across companies and jurisdictions.

\end{abstract}

\begin{figure*}[h!]
    \centering
    \includegraphics[width=\linewidth]{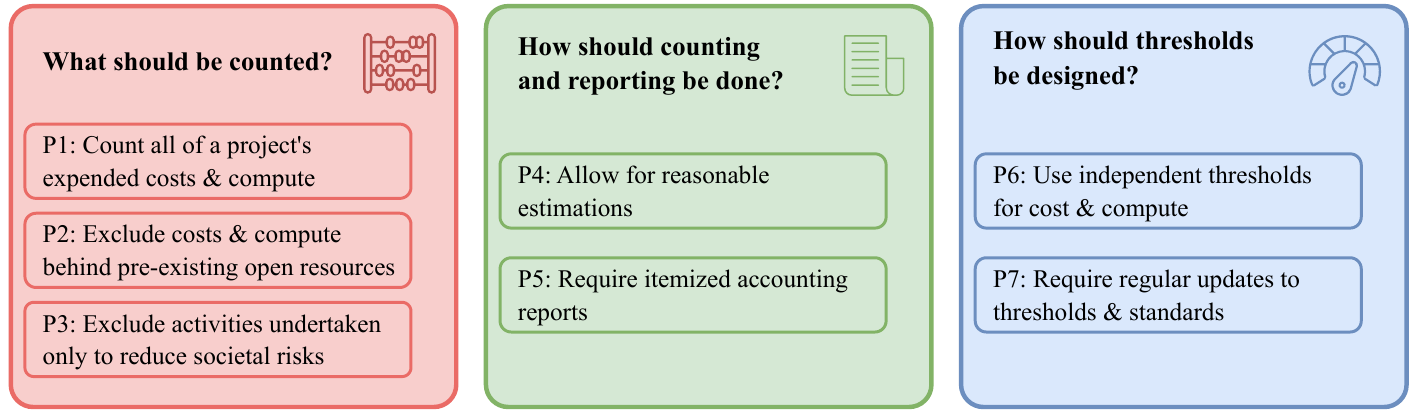}
    \vspace{-20pt}
    \caption{\textbf{Seven principles for cost and compute accounting discussed in \Cref{sec:principles}.}}
    \label{fig:principles}
\end{figure*}

\section{Introduction} \label{sec:intro}

As AI systems become more capable, policymakers face a mounting challenge: identifying which AI systems warrant heightened oversight. Recent laws, draft laws, and governance frameworks have approached this challenge by proposing requirements contingent on development costs and computational resources (e.g., \citealp{EU_AI_Act_Article_51, CA_SB_1047_2023, cafrontier2025draft, framework_ai_diffusion_2025, NY_A6453_2025, CASB53_2025, MIHB2025_4668}). 
For example, Article 51 of the EU AI Act states ``A general-purpose AI model shall be presumed to have high impact capabilities...when the cumulative amount of computation used for its training measured in floating point operations is greater than $10^{25}$.''

Development costs and compute are compelling metrics for use in AI governance \citep{cafrontier2025draft} because they ``correlate with capabilities and risks, are quantifiable, can be measured early in the AI lifecycle, and can be verified by external actors'' \citep{heim2024training}. Cost and compute thresholds also allow regulators to focus oversight on the most advanced AI developers while avoiding unnecessary burdens on smaller developers. However, their practicality as regulatory tools depends on establishing clear accounting standards. This requires resolving several technical ambiguities about what should be counted \citep{hooker2024limitations, heim2024training, reuel2024open}. 

This paper asks the question: \textit{\textbf{How can the cost and compute used during model development be counted in a way that is practical, limits gameability, and avoids disincentivizing responsible risk management?}} Key challenges include determining which activities to count, establishing reporting requirements, and allowing for standards to adapt as technology evolves.


To address these challenges, we propose seven principles for designing practical AI cost and compute accounting standards. We argue that these principles can resolve technical ambiguities while aligning with public interest and enabling consistent implementation across companies and jurisdictions. While we do not take positions on specific thresholds or requirements, our framework provides a foundation for developing robust standards for AI cost and compute accounting.

\section{Background} \label{sec:background}

\textbf{Related work.} AI research has studied ``scaling laws'' demonstrating how AI model performance improves predictably with increased computational resources and data \citep{kaplan2020scaling, Villalobos2023ScalingLaws, sevilla2024training}. While these relationships provide a scientific basis for the theoretical value of cost and compute thresholds in AI regulation \citep{sastry2024computing}, researchers have identified important limitations and highlighted unsolved implementation challenges. Without a standardized methodology, complex technical questions about which activities to count and how to report such counts remain unresolved \citep{hooker2024limitations, heim2024training, yew2025red}.

\begin{figure*}[t!]
    \centering
    \includegraphics[width=0.8\linewidth]{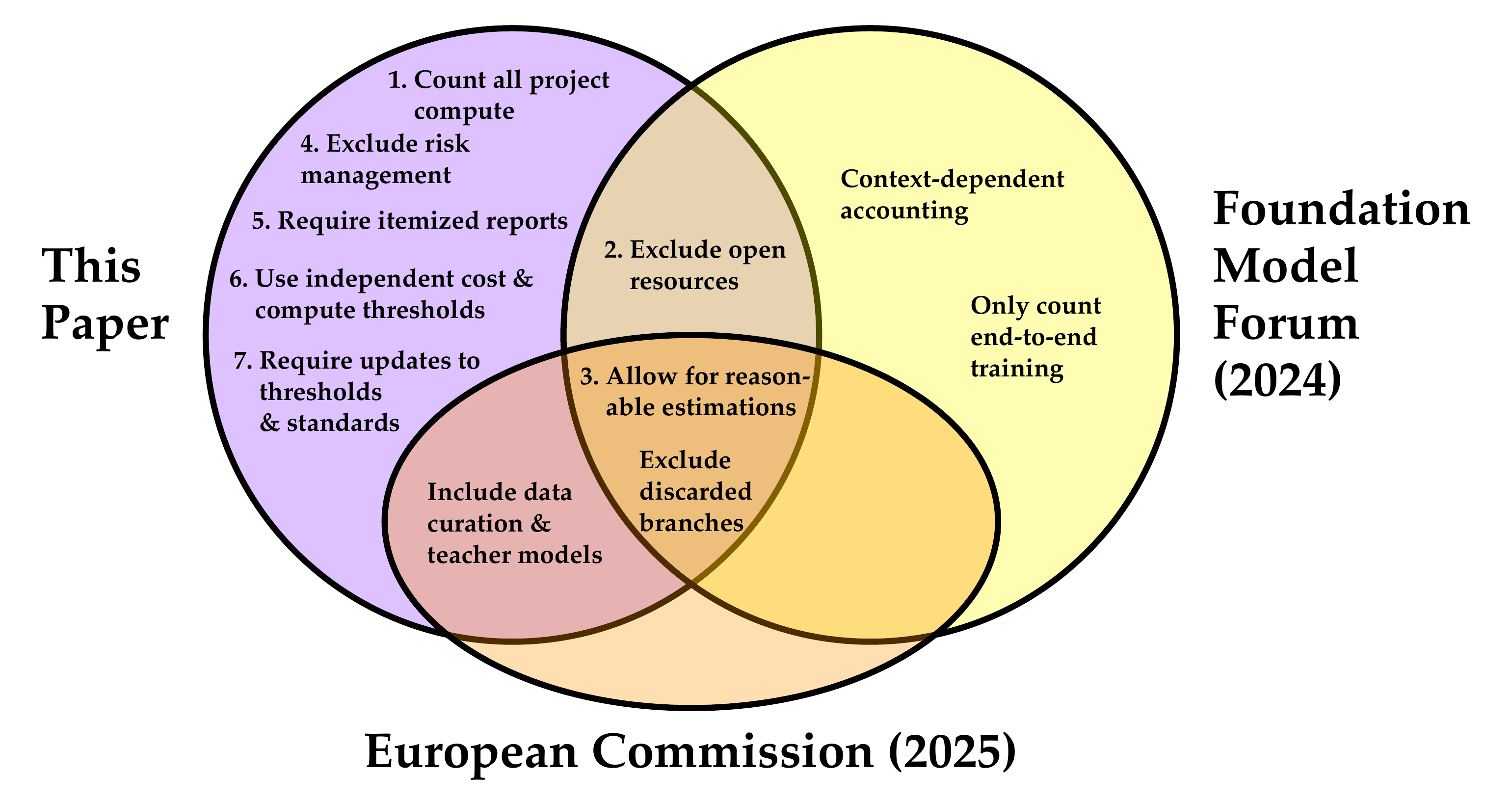}
    \caption{\textbf{Comparing principles proposed here, in \citet{FrontierModelForum2024MeasuringCompute}, and in \citet{europeancommission2025gpaiclarification}.} In \Cref{sec:principles}, we argue that some exclusions recommended by \citet{FrontierModelForum2024MeasuringCompute} and \citet{europeancommission2025gpaiclarification} could allow for loopholes. Most notably, the recommendations from \citet{FrontierModelForum2024MeasuringCompute} allow for the ``distillation loophole'' (see \Cref{sec:project}) in which developers sample-efficiently train a model on the outputs of a `teacher' model without accounting for the teacher's training.}
    \label{fig:venn_diagram}
\end{figure*}

To our knowledge, there are only two sets of open, pre-existing guidelines for cost and compute accounting. The first was published as an issue brief by the \citet{FrontierModelForum2024MeasuringCompute}, a collaboration between major tech companies that represents industry interests \citep{wei2024ai}, while the second appears in the draft guidelines on general-purpose AI issued by the \citet{europeancommission2025gpaiclarification} in support of the EU AI Act. 
In \Cref{fig:venn_diagram}, we compare the principles we recommend to those recommended by each of these prior reports.
Here, we echo the \cite{FrontierModelForum2024MeasuringCompute} in support of excluding open resources from accounting, the \citet{europeancommission2025gpaiclarification} in support of counting data curation and teacher models used for distillation, and both prior reports in their support of allowing for reasonable estimates. 
However, we argue that previously recommended practices, such as context-dependent calculations, limiting counts to end-to-end training only, and excluding recomputations and discarded branches, would enable developers to game the system by omitting substantial portions of their development process from oversight (\Cref{sec:project}).



\textbf{Key terms.} In the context of this paper:
\begin{itemize}
    \item \textbf{Developer} refers to the entity undertaking the process of creating an AI model. A single developer can encompass multiple legal entities in formal partnership.\footnote{We use this definition to preclude loopholes involving multiple legal entities formally collaborating to develop a single model. Detailed standards will need to account for collaborative developments, including crowdsourced or federated approaches.} 
    \item \textbf{Development} refers to the process of curating data, training models, creating scaffolding, and testing AI systems. It does not encompass human labor, operations, or procuring hardware.
    \item An \textbf{AI model} refers to a neural information processing structure trained using machine learning. An \textbf{AI system} refers to a set of one or more AI models combined with other software components to accomplish a specific task. For example, GPT-4o \citep{hurst2024gpt} is an AI model while ChatGPT-GPT-4o is an AI system. 
\end{itemize}

\textbf{What if multiple models are very similar?} It is possible to develop two distinct but very closely related models. For example, two models may only differ by a small amount of fine-tuning if they are different derivatives of the same `base' model. This poses a challenge to regulators because two such models will often, but not always, have similar behaviors. It may also often be impractical to subject multiple models to potentially redundant requirements. For this reason, policymakers may want to make developers or `model families' the object of regulation as opposed to individual models. However, recommendations for how to practically handle these cases are beyond the scope of this paper.

\section{Principles} \label{sec:principles}


\subsection{Count all of a project’s expended costs and compute upstream of the final system} \label{sec:project}

\textbf{Principle:} Count all technical costs and compute that the developer expends upstream of the final AI system, not simply theoretical or proximal ones. 

\textbf{Purpose:} Closing loopholes (especially involving distillation), and limiting the gameability of accounting standards. 

Developers undertake a variety of activities during frontier model development. However, a narrow view of what counts could be used to exclude certain activities integral to the model development process. For example:

\begin{itemize}
    \item Some activities are not theoretically needed for the final model to have been produced. For example, in the process of training models, there are often many multiplications or additions by zero due to the use of dropout \citep{srivastava2014dropout} and sparsity (e.g., \citealp{correia2019adaptively}). Even though they are not theoretically needed, they are carried out by hardware nonetheless and are often used to improve performance. 
    \item Some activities are not proximal to the model's training process. For example, dataset creation/curation/compression (e.g., \citealp{kaddour2023minipile, solaiman2021process, chen2024automated}) or training a teacher model for distillation \citep{yang2024survey} are not directly involved in the final model's training. However, these activities are nonetheless integral to the model’s development and capabilities. 
\end{itemize}

When there exists an incentive to make a model seem very cheap using narrow accounting standards, developers can find creative ways to game the count \citep{yew2025red}. 
One of the most pertinent ways that narrow accounting can be used to obscure a model's total development costs involves model distillation.
For instance, DeepSeek-V3 \citep{liu2024deepseek} was produced, in part, by distilling the more powerful DeepSeek-R1 \citep{guo2025deepseek}, yet its widely quoted \$6 million budget excludes the compute spent training DeepSeek-R1—making the total project appear far less costly than it was in actuality, demonstrating how narrow accounting can mask large hidden costs.
For this reason, accounting standards that do not close the distillation loophole in particular may be highly limited in their effectiveness.

\begin{quote}
    \textit{``Any statistical relationship will break down when used for policy purposes.''}\\
    -- Jón Danielsson 
\end{quote}


Finally, sometimes, there will exist genuine grey-areas related to whether a certain activity was a meaningful part of a model’s development process. For example, if a developer conducts basic research to develop techniques that they will use downstream for the development of a model, should that be counted? 
Since company R\&D processes are never completely isolated, some ambiguity in attributing activities to specific model development is inevitable. However, \Cref{sec:reports} will discuss reporting requirements as a mechanism for fostering transparency and accountability for cost and compute accounting practices.

\subsection{Exclude costs and compute behind pre-existing, openly-available resources} \label{sec:open}

\textbf{Principle:} Count costs and compute that the developer directly incurs through their activities, purchases, and partnerships. Exempt the costs and compute used to produce open resources that developers obtain for free. 

\textbf{Purpose:} Practicality and focus on proprietary resources.

Developers can produce capable models through multiple sources of cost and compute. They often curate their own data and train their own models in-house. However, they can also purchase resources, query systems from external providers, and outsource parts of the development process to partners. For the reasons outlined in \Cref{sec:project}, these are generally needed for thorough accounting.

We argue that the costs and compute behind pre-existing, openly-available resources should be excluded, both because their model- and data-provenance records are often inconsistent \citep{mitchell2019model,longpre2023data} and because such resources already provide a zero-effort capability floor.

However, one modification to this exemption may be necessary to close a loophole. If resources that were recently (e.g., within 6 months prior to when a model’s development begins) and openly released by the developer\footnote{Recall in \Cref{sec:background} that we define ``developer'' to include formal collaborations between multiple legal entities. This type of definition would prevent multiple legal entities from spliting the development process via open-weight checkpoints to avoid passing a threshold so long as they had a formal agreement to do so.} in question, regulators may wish to require that this resource is still counted. Without this exception, developers could openly release partially-developed model components (e.g., a pretrained base model) to exclude them from accounting. 

As a final note, regulators may wish to uniquely handle cases in which a developer begins with an open model whose development already passed thresholds and further develops it. For example, a failed 2024 California bill \citep{CA_SB_1047_2023} defined a ``covered'' model in terms of either a primary threshold or a secondary threshold for when additional development is applied to an existing ``covered'' model. This type of strategy may be appealing to regulators because of how modest amounts of further development of highly capable models can substantially alter their capabilities. However, recommendations on how regulators should handle these cases are beyond the scope of this paper.

\subsection{Exclude activities undertaken only to reduce societal risks} \label{sec:safety}

\textbf{Principle:} Allow developers to exempt activities undertaken only for the purpose of reducing risks to society which do not have side effects of enhancing model capabilities.  

\textbf{Purpose:} Incentivizing societal risk-reduction practices.

Over the course of developing an AI model, key activities are undertaken either partially or entirely to improve its capabilities. For example, pretraining and fine-tuning are principally meant to make models more capable. However, some activities are undertaken strictly to reduce risks. Examples include filtering child sexual abuse material (CSAM) from training data \citep{thiel2023identifying}), fine-tuning models to refuse criminal requests \citep{yuan2024refuse}, and testing for national security risks \citep{shevlane2023model}. To avoid disincentivizing such measures, developers must be allowed to exclude these types of activities from their accounting. 

How should it be determined when an activity is undertaken only for the purpose of mitigating societal risks? This can be difficult due to the lack of a clean dichotomy and the prevalence of ``safetywashing'' \citep{ren2024safetywashing}. To mitigate this challenge, developers can be required to produce a rigorous, auditable justification for why an activity only reduces risks without simultaneously increasing capabilities in an accounting report (see \Cref{sec:reports}).

\subsection{Allow for reasonable estimates} \label{sec:estimates}

\textbf{Principle:} When counting costs and compute used for a model’s development, developers should be permitted (and often expected) to use reasonable estimations when precise information is not practically attainable. 

\textbf{Purpose:} Practicality.

Information about costs and compute expended during a model’s development is not always precisely quantifiable. For example, developers will often not know exactly how much compute has been expended when they query a closed-source system from some outside provider. However, in a case like this, reasonable estimates can be made based on contextual knowledge and the market value of compute \citep{sevilla2022estimating, cottier2024rising}. 
Such estimates parallel `fair value' asset estimations in financial accounting (IFRSF, 2022). Some imprecision is inevitable. However, to reduce the risk of estimations being gamed or resulting in unreliable counts, developers can be required to provide a report on their approach to accounting that documents estimates and justifications (see \Cref{sec:reports}). Meanwhile, regulators or standards bodies could publish guidance on appropriate estimation methodologies, tolerable error margins, and suitable documentation templates.

 \subsection{Require itemized accounting reports} \label{sec:reports}

\textbf{Principle:} Require developers to produce an auditable, itemized accounting report detailing their approach to accounting, including justifications for estimates and exemptions. 

\textbf{Purpose:} Transparency and accountability.

In financial accounting, companies are regularly required to send records and reports to governing bodies (e.g., \citealp{SEC_FRM_Topic_2}). This has both the direct effect of helping government oversight offices spot issues and the indirect effect of incentivizing due diligence from companies. The same applies to AI cost and compute accounting. These reports would also be key for developers to provide explanations and justifications for technical exemptions (\Cref{sec:project}), open resource exemptions (\Cref{sec:open}), risk mitigation exemptions (\Cref{sec:safety}), and estimations (\Cref{sec:estimates}). Such reports would improve accountability around accounting practices and inform regulators about industry trends in development expenditures.

For accounting reports to be effective, they must contain sufficient detail to enable meaningful review. 
For example, regulators may wish to offer standardized guidance on reporting including, for each distinct activity involving data curation, pretraining, fine-tuning, or testing:
\begin{itemize}
    \item A clear description of the activity and its purpose.
    \item An explanation (and, if necessary, evidence) for if and how the activity relied on open resources.
    \item An explanation (and, if necessary, evidence) for whether the activity was undertaken \textit{solely} for risk reduction.
    \item An explanation (and, if necessary, evidence) for estimations involved in accounting for that activity.
    \item The calculation used to quantify cost and compute for the activity. 
    \item The activity's final accounted cost and compute. 
\end{itemize}



\subsection{Use independent thresholds for costs and compute} \label{sec:independent}

\textbf{Principle:} Regulatory requirements should be independently triggered by separate thresholds for cost and compute. 

\textbf{Purpose:} Closing loopholes, and limiting the gameability of accounting standards.

Cost and compute are correlated, but they can still be decoupled, especially when developers have an incentive to game standards. For example, machine-generated data is cheap but computationally intensive while human-generated data is expensive but computationally free. A developer could design a project to be low-cost/high-compute or vice versa by adjusting the extent to which they use machine- versus human-generated data. As a result, having both cost and compute triggers would reduce gameability. 

Separate cost and compute thresholds can also serve as failsafes for each other in case of error or fraud. For example, if a developer purchases queries or other services from an external provider, precise information on the amount of compute used might not be available, but the costs are unambiguous and auditable. Meanwhile, different computing devices can use different amounts of power to perform the same computations, but the compute is unambiguous.

\subsection{Require regular updates to thresholds and standards} \label{sec:updates}

\textbf{Principle:} Require that thresholds and accounting standards are regularly updated to reflect technological developments. 

\textbf{Purpose:} Ensuring standards remain effective by adapting to technological advances and evolving societal needs.

Rapid developments in AI technology create uncertainty about how scaling trends and efficiency gains will affect developers' costs and compute requirements \citep{pilz2023increased}. Accordingly, governance frameworks will need to be adaptive to ensure they remain relevant over time. To regulate incisively, government offices and/or standards bodies will need to revisit and curate standards on a regular (e.g., quarterly or semiannual) basis in response to new developments in the state-of-the-art.

\section{Discussion}

\textbf{Significance:} 
Regulatory thresholds involving cost and compute are a uniquely practical \citep{heim2024training, cafrontier2025draft} yet technically challenging \citep{hooker2024limitations} strategy for designing regulations that target frontier AI models. 
To make cost and compute thresholds more tenable as a regulatory strategy, standards for accounting must be clear, consistent, and aligned with public interest. To support the development of such standards, we have proposed a principles-first framework to resolve ambiguities and introduced seven principles designed to reduce gameability, avoid disincentivizing societal risk mitigation practices, and enable consistent implementation across companies and jurisdictions. 


\textbf{Drawbacks:} Despite the benefits discussed in this paper, rigorous accounting standards can also come with drawbacks. These may including undesired regulatory burden and information security risks that come from sharing of potentially sensitive information with regulators. Regulators should weigh these potential drawbacks against the benefits discussed above when designing standards.

\textbf{Limitations:} This work was not written in the context of any specific law. We make no recommendations about what kinds of regulatory requirements should be triggered and how. Key questions about how high to set thresholds, what they should trigger, and how they should be incorporated into legal frameworks are all beyond the scope of this paper. Furthermore, as we have discussed, while the principles presented here can greatly reduce ambiguity, grey areas will be inevitable. However, this underscores the role that itemized reporting can play in regulatory awareness. Due to the constantly evolving nature of AI technology, we recommend a paradigm of maintaining continuous awareness and oversight while allowing regulatory regimes to evolve rather than pursuing a single fixed regime. 

\textbf{Future work:} Whereas this paper has sought to outline principles for designing standards, future work will be needed to produce concrete standards. Implementing these principles in practice will require specific attention to the purpose and scope of any individual law and may require compromises to ensure logistical and/or political feasibility.

\newpage
\section*{Acknowledgments}

We would like to thank Anthony Aguirre, Alexander Erben, James Petrie, Joe Kwon, Lennart Heim, Mark Brakel, Nandi Schoots, and Richard Mallah for discussions and feedback. 
Stephen Casper and Luke Bailey are funded by a Future of Life Institute Vitalik Buterin Fellowship. Luke Bailey is also funded by the SAP Stanford Graduate Fellowship.

\bibliography{bibliography}
\bibliographystyle{icml2025}





\end{document}